\documentclass[runningheads]{llncs}
\usepackage[T1]{fontenc}
\usepackage{latexsym}
\usepackage{color}
\usepackage{amsfonts}
\usepackage{multirow}
\usepackage{color, colortbl}
\usepackage{graphicx}
\usepackage{amsmath}
\usepackage{xurl}
\usepackage{caption}
\begin{document}
\title{Deep Neural Network for Constraint Acquisition through Tailored Loss Function}
%
%
\author{Eduardo Vyhmeister \inst{1}\orcidID{0000-0003-1922-2706} \and
Rocio Paez\inst{1}\orcidID{0000-0001-5885-4787} \and
Gabriel Gonzalez\inst{1}\orcidID{0000-0003-0486-1492}}
\authorrunning{F. Author et al.}
%
\institute{Insight Centre of Data Analytics, University College Cork, Cork, Ireland \email{eduardo.vyhmeister@insight-centre.org}}
\maketitle              
\begin{abstract}
The significance of learning constraints from data is underscored by its potential applications in real-world problem-solving. While constraints are popular for modeling and solving, the approaches to learning constraints from data remain relatively scarce. Furthermore, the intricate task of modeling demands expertise and is prone to errors, thus constraint acquisition methods offer a solution by automating this process through learnt constraints from examples or behaviours of solutions and non-solutions. This work introduces a novel approach grounded in Deep Neural Network (DNN) based on Symbolic Regression that, by setting suitable loss functions, constraints can be extracted directly from datasets. Using the present approach, direct formulation of constraints was achieved. 
Furthermore, given the broad pre-developed architectures and functionalities of DNN, connections and extensions with other frameworks could be foreseen.


\end{abstract}

\section{Introduction}

The importance of learning constraints from data is emphasized by its potential applications in addressing real-world problem-solving scenarios. Despite the widespread use of constraints for modeling and problem-solving, the methodologies for learning constraints from data are relatively limited. Recent efforts have been directed towards surveying and contextualizing constraint learning within the broader machine learning landscape. This involves recognizing subtle distinctions from standard function learning, identifying challenges, and exploring potential applications of constraint learning in diverse fields. These endeavors contribute to a deeper understanding of the role of constraint learning and pave the way for its integration into a wider array of problem-solving domains.

In terms of existing approaches for constraint acquisition, various systems (covered in the following section) have shown capabilities to extract systems constraints using iterative methods, oracle feedback, examples and constraints catalogues, to reduce the constraint space. Most of these deductive methods have different limitations that restrict broad applicability. For example, the dependence on user interaction may pose challenges, especially if users are not domain experts or if providing examples requires a deep understanding of the problem. The number of queries or interactions needed to converge to an accurate model can also be a limitation.  Furthermore, 
for some methods based on exemplication, a priori knowledge of the positive 
or negative nature of the examples is 
required to limit the search spaces, which implies the need for previous knowledge of the system behaviour. Additionally, the effectiveness of these methods relies on the expressiveness of the predefined constraint libraries or the set of possible constraints. If the library lacks certain types of constraints relevant to a specific problem domain, the models generated may be limited in their accuracy. 

In the broader context of constraint acquisition, the interplay among machine learning, data mining, and constraint satisfaction has attracted considerable attention. While constraints are conventionally employed in logical and optimization problems, their application in the domains of machine learning and data mining is becoming increasingly prominent (e.g \cite{donato_2023}). 

The term "constraint learning" is precisely defined in this context as the inductive learning of a constraint theory from examples. This definition distinguishes constraint learning from the deductive processes associated with clause learning or constraint aquisition of non-ML solver technologies. The exploration of constraints in the context of machine learning and data mining signifies a multifaceted and evolving area of research with implications for diverse problem-solving domains.

The integration of Deep Neural Networks (DNN) into constraint learning brings benefits, leveraging their ability to efficiently extract information directly from data. DNN can learn intricate representations, discover relevant features, and adapt to different data types. 
DNN excel at learning intricate and hierarchical representations from large data sets \cite{Bau_2020}. They can automatically discover relevant features and patterns in the input, which can be advantageous in capturing complex relationships within constraint networks. Furthermore, Deep learning models can be trained in an end-to-end fashion, allowing them to learn both the feature representation and the constraint-solving process simultaneously. This may lead to more seamless integration and optimization. Additionally, DNN are highly adaptable to different types of data and problem domains \cite{Donges_2023}. They can be trained to handle diverse constraint types and can potentially generalize well to new problem instances. DNN can automatically extract relevant features from the input data, potentially reducing the need for manual feature engineering in the modeling process. This can be particularly beneficial when dealing with complex and high-dimensional constraint spaces, nevertheless that do not imply that expert knowledge could not be integrated within DNN methodologies \cite{dash_2022,sun_2020}. In fact, active learning can be naturally integrated into neural network-based approaches. The model can dynamically query the user or another system for specific examples, enabling more efficient and targeted learning of constraints. Deep learning models can take advantage of parallel processing capabilities, which can lead, together with increasing computational power, to faster training times \cite{Donges_2023}. This can be crucial in scenarios where quick acquisition of constraint knowledge is essential. Furthermore, pre-trained neural network models can be fine-tuned for specific constraint learning tasks \cite{Church_2021}. This leverages knowledge learned from one context to another, potentially speeding up the learning process for similar new problems. Importantly, Deep learning techniques can be seamlessly integrated with other machine learning methods or other pre-trained DNN, providing a holistic approach to constraint learning. This can include combining neural networks with symbolic reasoning or other constraint-solving techniques.

While these benefits are promising, it is important to note that the success of neural network-driven approaches in constraint learning depends on factors such as data availability, problem complexity, and the specific characteristics of the application domain. Additionally, interpretability and transparency of neural network models should be considered, especially in domains where understanding the reasoning behind constraint decisions is crucial.

Symbolic regression (SR) is a computational method employed to uncover mathematical equations, commonly implemented through genetic programming. This entails using evolutionary algorithms to explore and identify the most suitable equation that aligns with a given dataset. While SR has proven effective in revealing fundamental laws governing physical systems based on empirical data, its scalability is hindered by the combinatorial complexity of the underlying problem \cite{he_2022} \cite{srbench_2022}.

The synergy between SR and DNN is pivotal in the pursuit of identifying behaviours inherent in a dataset. SR operates by exploring mathematical expressions to discover the optimal model that aligns with the given data. On the other hand, Neural Networks (NN) and Machine Learning (ML) techniques excel in capturing intricate non-linear relationships between input and output variables.

For instance, Schmidt et al. \cite{schmidt2009distilling} and Martius et al. \cite{martius2016extrapolation} introduced a DNN approach named the Equation Learner (EQL), where traditional activation functions are replaced with primitive functions (PFs). This novel approach enables the DNN to perform SR, allowing it to learn analytical expressions and extrapolate to unseen domains. The EQL is implemented as an end-to-end differentiable feed-forward network, facilitating efficient gradient-based training. This integration of SR and DNN methodologies represents a powerful strategy for uncovering and understanding complex relationships within datasets.

To evaluate the potential of DNN for constraint acquisition, with all the possible benefits previously described, and, at the same time, facilitate the transparency of the DNN for improved user acceptance of the final results, this work proposes the use of an EQL-based architecture with tailored loss functions to extract system constraints by using tailored loss function. The manuscript is organized as follow: In section \ref{relatedworks} are presented recognized systems for constraint acquisition. In Section \ref{methodology} an introduction to the EQL, the tailored loss function, and their combination, are described. This section also presents the experiments used for the system evaluation. Section \ref{results} discusses the results. Finally, 
Section \ref{conclusion} presents the main remarks and conclusions. 

\section{Related Works} \label{relatedworks}

Constraint programming involves both modeling and solving. In modeling, problems are defined using variables with specific values and rules, while in solving, values are identified that satisfy all constraints simultaneously. Despite its potential, user-friendliness limitations pose challenges for constraint programming. The declarative nature of this approach enables the solving of problem models using standard methods, but fully leveraging this potential requires increased automation in modeling \cite{barry_2010}.

Constraint acquisition plays a crucial role in the automation of complex and error-prone tasks associated with modeling constraint programming problems. By using automated approaches constraints not previously considered or improvements on mathematical representations of those already considered could be achieved, allowing at the end to have better models to represent the system behaviour. The constraint acquisition role increase is important especially when the number of features and their trends complexity is increased. This process enables the extraction of constraints from data that represents both solutions and non-solutions. Various constraint acquisition systems have been proposed to support non-expert users in their modeling tasks \cite{Daoudi_2016}.

As an illustrative example, Baldicenau and Simonis \cite{Beldiceanu_2016} have introduced a methodology where constraints are constructed from a catalog as primitives. Their ModelSeeker tool leverages this approach to analyze a substantial dataset with up to 6500 variables and 7000 samples. In this methodology, examples are organized as a matrix, and the system identifies constraints in the global constraint catalog that are satisfied by rows or columns across all examples. An integral part of their workflow includes dominance checks and trivia removal, processes that assess constraints between each other or utilize specific rules to simplify and eliminate irrelevant constraints. These last steps can contribute significantly to enhancing the efficiency and relevance of the acquired constraints but, it is noteworthy that, as the system's complexity increases, users are required to provide a larger number of examples for the target set of constraints to be effectively learned, limiting the applicability of the approach to highly complex systems and feedback from expert knowledge.

Bessiere et al. \cite{Bessiere_2015} propose QuAcq, an iterative method generating partial queries and utilizing oracle feedback to reduce the constraint space. The need to use Oracle feedback imposes over QuAcq (and other approaches based on this) similar limitations discussed AS for Modelseeker.  Furthermore, the approach could also suffers from a high query count for convergence. MultiAcq \cite{Areangioli_2016} extends QuAcq, learning multiple constraints on negative examples. MQuAcq \cite{Tsouros_2020} combines QuAcq and MultiAcq, effectively reducing query complexity. 

Other approaches, such as PREDICT\&ASK introduce a distinct algorithm focused on predicting missing constraints within a partially learned network \cite{Daoudi_2016}. The algorithm uses the local data structure which contains all constraints that are candidate for recommendation. Through recommendation queries, the approach enhances user interaction, as demonstrated through experimental comparisons against QuAcq. Even though the recommendation approach could enhance system performance, the need for expert knowledge within the identification process force to have similar limitations as previously discussed systems.

Hassle-sls integrates metaheuristic techniques for joint learning of hard and weighted soft constraints. Despite advancements, runtime issues arise from evaluating multiple MAX-SAT models. Hassle-gen \cite{berden_2022} addresses this by incorporating a genetic algorithm and an efficient model evaluation procedure, contributing significantly to the state of the art.

\section{Methodology}\label{methodology}

The general goal of the approach is to find expressions of the form $-B + \sum_{i=1}^{F} A_i \cdot f(X_i)$ that delimit (i.e. constraints) the space in which the dataset is distributed. In this expression $A$ and $B$ are the corresponding terms of the constant matrices to be solved during the training process and $F$ is the number of features in the dataset. It is important to note that while $f(X_i)$ is represented as a linear expression in this work, it has the flexibility to assume any form by setting sound primitives in the EQL. For a better understanding of the EQL and its interaction with the tailored loss functions, we first include a short description of SR.

\subsection{Symbolic Regression}

In the context of SR, the goal is to model a system represented by an unknown analytical function, denoted as $\phi: \mathbb{R}^n \rightarrow \mathbb{R}^m$. The observed data, denoted as $x, y = {(x_1,y_1),\ldots,(x_N,y_N)}$, where $x \in \mathbb{R}^n$ and $y \in \mathbb{R}^m$, is generated from $y = \phi(x) + \xi$, where $\xi$ represents a term of additive zero-mean noise. The goal of SR methods is to construct a function $\psi: \mathbb{R}^n \rightarrow \mathbb{R}^m$ that minimizes the empirical error on the training set and generalizes well to future data.

Given that the true analytical function $\phi$ is unknown, the resulting mathematical expression, denoted as $\psi$, is designed to be interpretable and capable of extrapolating the data. This interpretability stems from the fact that the constructed model is not a black-box, making it easier to understand and analyze. The process involves considering a predefined set of "primitive functions," denoted as $\mathbf{f} = { f_1, f_2, \ldots, f_n}$, upon which the regression method can build. The selection of these primitive functions is crucial for accurately capturing the mathematical representation of the underlying model.

For instance, common choices for the set of primitive functions, denoted as $\mathbf{f}$, may include constants ($C$), linear terms ($x$), quadratic terms ($x^2$), trigonometric functions ($\sin(x)$), exponential functions ($\mathrm{e}^{x}$), sigmoid functions ($S(x) = 1/(1+\mathrm{e}^{-x})$), logarithmic functions ($\mathrm{ln}(x)$), reciprocal functions ($1/x$), and square root functions ($\sqrt{x}$). The specific choice of primitive functions can be tailored to different applications and problem domains. In the present work to construct linear constraints, primitives were limited to linear terms and constants.

Specification of the neural architecture, together with the integration with the specific loss function used for constraint acquisition is described in Section \ref{integration}

\subsection{Loss Function Definition for Constraint Acquisition}




ML and DNN algorithms can be fine-tuned or trained by solving a problem-structural optimization, expressed as \cite{wang_2022}:

\begin{equation}
    \min_{f} \frac{1}{N} \left[ \sum_{i=1}^{N} L_{\theta}(f(x_i)) + \lambda R(f) \right]
\end{equation}

In this minimization problem, a training set of size $N$ is assessed through two main terms. The first term is the empirical risk, where $L()$ denotes the loss function, $\theta$ represents the parameter vector, and $f$ is a functional evaluation over each element $i$ of the set. The second term, $R(f)$, is known as the regularization term, reflecting the model's complexity. Including this second term aids in reducing the size of the final expressions. The weight parameter $\lambda \geq 0$ balances the trade-off between the empirical risk and the model complexity. Moreover, additional terms can be introduced and appropriately weighted to extend functionalities during training.

Common regression loss functions like square loss, absolute loss, Huber loss, and Log-cosh loss are designed for deriving functions in regression tasks (e.g., $f(x) = \phi(x) + \xi$ from the dataset). In constraint learning, where expressions involve both equalities and inequalities, modifying the loss function is crucial. This adaptation emphasizes adherence to specified constraints, distinct from conventional regression that mainly minimizes disparities between model predictions and observed values. The redefined loss function guides the model to respect given constraints, shifting focus from precise numerical predictions to a more comprehensive consideration of data limitations.

To address this need, it is proposed the utilization of a tailored loss function,
taking into account three inter-playing notions:

1. In our approach, we track the permissible directions for adjusting the predicted function by employing two distinct expressions of error. The choice between these expressions is contingent on the type of inequality sought. For instance, if the objective is to enforce constraints of the form $f(x) \leq A$, where $A$ is a specified parameter, the error expression $y - f(x)$ is employed. Conversely, when aiming for constraints of the form $A \leq f(x)$, the opposite expression, $f(x) - y$, is utilized. Each of these error terms is weighted to contribute to an overarching global valuation term.

2.
The previous term is combined with with an absolute maximum error that serves as an anchor to bound the extent of movement in the predicted equation space. The inclusion of this anchor ensures that the optimization process does not diverge infinitely, promoting stability in the learning process. 

3. Finally, an induced observation threshold derived from the spectrum of all possible errors is considered; rather than taking into account errors over the entire dataset. 

\begin{figure*}\label{fig1}
\centering
\includegraphics[scale=0.8]{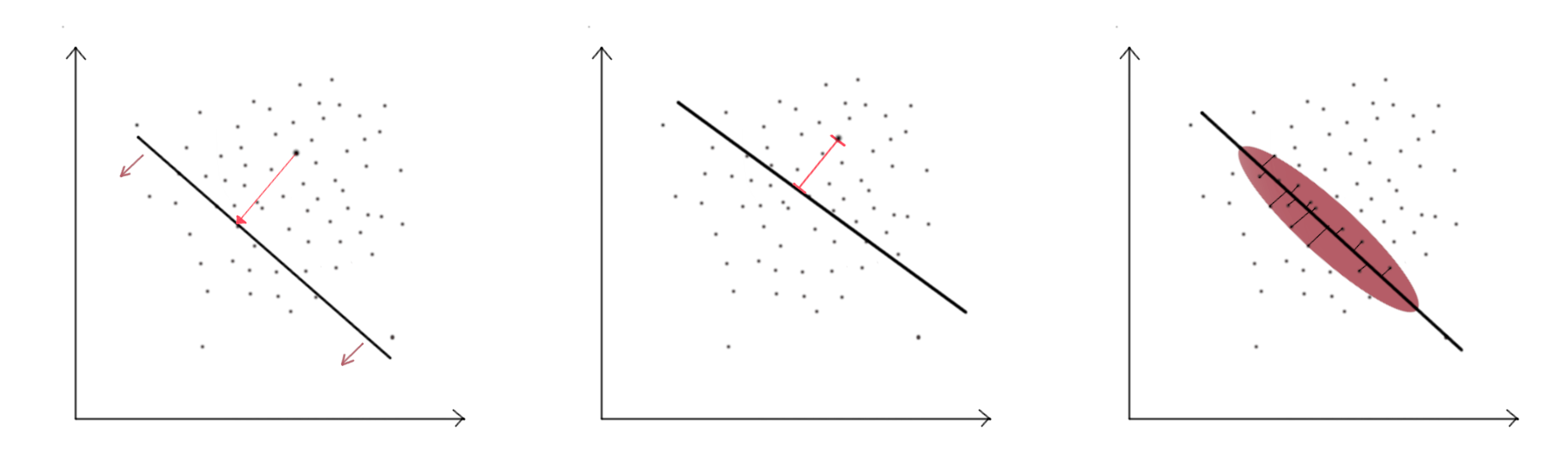}
\captionsetup{belowskip=0pt}
\caption{Schematic representation of the different notions considered for the three terms definition the loss functions: Notion 1 (left panel), Notion 2 (center panel), Notion 3 (right panel). See text for full explanation.} \label{fig_schema}

\end{figure*}

This approach combining these three notions effectively constrains the model's behaviour, aligning with the overarching goal of capturing constraints inherent in the data. The graphical interpretation of these notions are shown in Figure \ref{fig_schema}.

These notions are expressed within the loss function mathematical representation
as three distinct and independent terms, 
implemented in the tuning (minimization) process. These components are carefully designed to accommodate the specific nature of the constraints being searched:

\begin{equation}\label{eq3}
    z = \frac{\alpha_1}{N}  \sum_{i=1}^{N} L_{e}(y,f(x_i)) + \frac{\alpha_2}{N}  \sum_{i=1}^{N} L_{P_{alpha}}(y,f(x_i)) + \alpha_3  L_{anchor}(y,f(x_i))  \\
\end{equation}
\begin{equation}\label{eq4}
    e =
    \begin{cases} 
    y  - f(x)  \text{, if } A \leq f(x) \\
    f(x) - y  \text{, if } f(x) \leq A 
    \end{cases}
\end{equation}
\begin{equation}\label{eq5}
    L_{e}(y,f(x_i)) = e
\end{equation}
\begin{equation}\label{eq6}
    L_{P_{\gamma}}(y,f(x_i)) =
    \begin{cases}
        (y_i - f(x_i))^2 \; \; \forall \; \text{y,xi if error} \in P_{\gamma} \\
        0 \; \; \forall \; \text{y,xi if error} \notin P_{\gamma} \\
    \end{cases}
\end{equation}
\begin{equation}\label{eq7}
    L_{anchor} (y,f(x_i)) = | Max(e) | \\
\end{equation}






In these equations, the minimization objective is denoted as $z$, and $\alpha$ represents the weighting factor assigned to each contribution during the training. The set $P_{\gamma}$ comprises data points within the $\gamma$ percentile.

The term $L_{\text{error}}$ captures the directional aspect of the loss, contingent on the inequality direction (notion 1). For instance, choosing the error  $ e = y - f(x)$ promotes higher negative values as minimization is carried out, favouring constraints of the form $A \leq f(x)$.

On the other hand $L_{P_{\gamma}}$ represents a quadratic loss function, with the distinctive feature that it considers only data points within the $\gamma$ quartile (notion 3). This focused observation facilitates obtaining robust representations of constraints, particularly in border regions identified during the minimization process. Quantil-based analyses and loss functions have found applications in various fields such as statistics and econometrics, underscoring its versatility and effectiveness in capturing the range of potential outcomes. This adaptability makes it a promising choice for tasks involving constraint acquisition, where a broader consideration of prediction intervals is warranted.



Lastly, $L_{\text{anchor}}$,is the absolute value of the maximum error (notion 2). Unlike $L_{\text{e}}$, this term acts as a counterbalance by progressively increasing and anchoring $f(x)$ within the proximity of the data points. It prevents the model from diverging infinitely from the observed data points.


\subsection{EQL and Loss Function Integration}\label{integration}

The integration of both SR and the custom loss function for constraint acquisition was done using Python as the primary programming language. TensorFlow V2 served as the framework for constructing the EQL, with activation functions explicitly specified as primitive functions defined by the user. To facilitate a direct understanding of the mathematical constraint representation and potentially offer feedback to the symbolic DNN, the \textit{sympy} library was employed as a symbolic mathematical tool.


By leveraging Tensorflow's capabilities, the custom loss function was incorporated during the compilation process of the optimizer. The full implementation of the 
combined DNN architechture can be found in the associated repository (\url{https://github.com/eduardovyhmeister/Constraint-Aquisition}), including symbolic layers (tensorflow) and methods such as initialization, build, call, and configuration. These symbolic layers facilitate the integration of symbolic mathematics into the DNN, enhancing interpretability and enabling feedback mechanisms for constraint representation.


For the experiments that follow, a three-layer architecture comprising an input layer, a symbolic layer, and an output layer was used. The number of layers can be varied for serving different goals. The input layer allows the data from a varying number of features to input the system. The symbolic layer allows defining which primitives functions are active (and parametrized during training). This layer used the identity (i.e., $f(x)=x_i$) and constants ($C$) as primitive without any bias on the neurones, suitable for linear constraints. The output layer, which consists of a single node representing the output prediction, aggregates the evaluations on the symbolic layer using a linear activation function (that includes a bias). This layers also included L1 and L2 regularization for training considerations. 

Additionally, a masking process was integrated during training in some the 
experiments. If a weight parameter fell below 0.001, the connection weight value was 
set to 0 and could not be updated thereafter. This approach promotes the parsimony 
of the final equation by eliminating any contribution from connections with 
negligible impact.

In the Equation Learner, the values of the weights and biases in the architecture play a critical role in defining the mathematical formulation. Thus, a careful initialization of these values is essential to effectively explore the search space during the training process. For the symbolic layer, Xavier uniform initializers were used, while random uniform initializers were employed for the biases of the output layer. The random uniform initialization was confined within 0.5 times the minimum and 0.5 times the maximum of the provided data values. It is important to note that the analysis of optimal initializers for the search engine was not explicitly considered in the current work, leaving room for further investigation in future studies.

To extract the mathematical formulation from the Equation Learner (EQL), the input data should include each of the features considered in the analysis (i.e., $X_i$). The output was defined as an array of zeros with the same size as the instances under evaluation.


\subsection{Experiments definition and setup}
Table \ref{tab1} shows the metaparameters to configure both, the EQL (first two columns) and the loss function (last two columns). 

\begin{center}
\begin{table}
\caption{Parameters for EQL and the loss function}\label{tab1}
\centering
\begin{tabular}{|c|c||c|c|}
\hline
Metaparameter &  Value & Metaparamter & Value\\
\hline
 Activation Functions & two $f(x)=x_i$ and two $C$ & $\alpha_1$ & 1.0 \\
 Epochs & 400  & $\alpha_2$ & 0.5 \\
 Learning rage & 1e-8  & $\alpha_3$ & 0.5 \\
  L1 & 0.05  & $\gamma$ & 5.0 \\
   L2 & 0.05  & & \\
\hline
\end{tabular}
\end{table}
\end{center}

Testing the approach and the loss function involved creating data points randomly based on a specific set of constraints. For a two-dimensional problem defined 
 by $X_0$,$X_1$, the 
 experiments include data with the following configurations: 
 \begin{itemize}
\item[1-] High Granularity Square: 
 an area with 600 data points delimited within -5 and 25 with an additional restriction (i.e. $ -5 \leq X_i \leq 25$, $i=0,1$ ; $4 < X_1+2X_2$)
 \item[2-] Circle: a circular area of 
radius $\sqrt{200}$ containing 250 data points (i.e. $X_{0}^2+X_{1}^2 \leq 200$); \item[3-] Low Granularity Square: 
same area as (1) with only 100 data points.
\end{itemize}
For a three-dimensional case study 
(defined by $X_0$, $X_1$, $X_2$), 2000 data points were distributed in a cube delimited within -5 and 25 with a planar section (i.e.  $-5 \leq X_i \leq 25$, 
$i=0,1,2$;   $4 < X_1+2X_2-3*X_3$). 

The data set, a combination of input ($X_i$) and output $Y$ data points within the given constraints, was fed to the EQL with the tailored loss function. The training process was run 10 times using a specific error representation.


\section{Results and Discussion} \label{results}


Figure \ref{fig1} illustrates the results when two features and only one type of error was used (i.e. $y-f(x)$). 

\begin{figure*}\label{fig1}
\centering
\includegraphics[scale= 1.1]{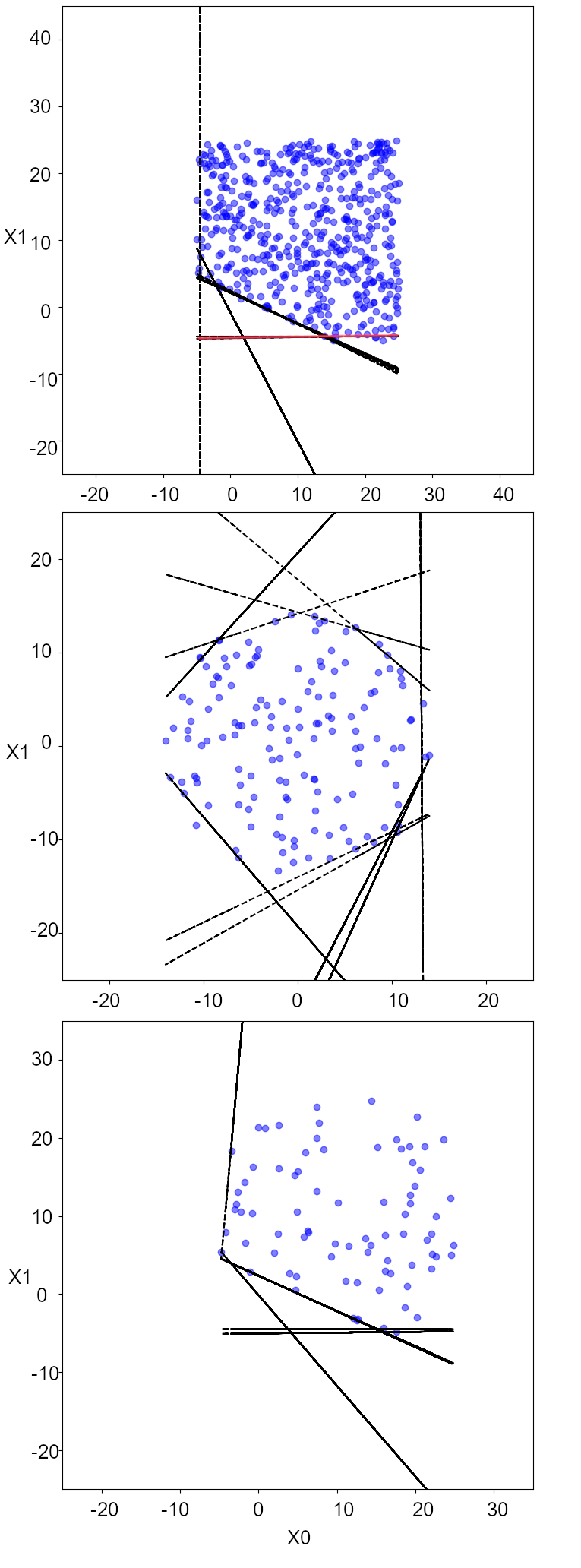}
\captionsetup{belowskip=0pt}
\caption{Constraints obtained using $y  - f(x)$ representation of error within the $L_{e}$ expression: High-granularity Square example (top panel), Circle example (center panel), Low-granularity square example (bottom panel). Red line is equivalent to Result $\# 7$ of Table \ref{tab2}, without masking (see text for full explanation).} \label{fig1}

\end{figure*}

As observed in the figures, the EQL combined with tailored loss function was able to be used to extract suitable constraints for the data sets, independent of the figure shape (Square and Circle) or the granularity of the data points (600 or 100 for Square data points). 
As the granularity was reduced, the percentile value $\gamma$ had to be reduced from 5.0 to 2.5 to improve performance. As expected, for low granularity, 
the points within the same percentile of error are more separated, and a $\gamma$ value suitable for high granularity would imply points too distant from the boundary of the figure at low granularity. 
Thus, limiting the $L_P$ by reducing the $\gamma$ value contributes to a more accurate search.  Thus, $\gamma$ plays a crucial role depending on the system granularity.

As defined in the methodology, the directional expression of the error was specified as 
$y  - f(x)$, implying that constraints of the form $A \leq f(x)$ should be obtained. Yet, One important outcome observed during experiments with the Circle is that constraints of the form $A \leq f(x)$ and $f(x) \leq A$ were obtained. 
This is probably due to the fact that no constraints on the 
network weights were defined and thus, possible negative values for 
those weights were obtained (inversion of the constraint). 


Table \ref{tab2} display the results of the constraints found for all the
experiments performed, including the specific functionality detected in each 
case and the performance metric. The table 
includes the results for the High Granularity Square (Results from 1 to 10); Cube (results from 11 to 15); Low Granularity Square (results from 16 to 25); and Circle (Results from 26 to 35). The performance presented in the table is measured as the error quantified by the percentage of points violating the obtained constraints.
This error, when approaching a value close to 0 (though not necessarily reaching 0), implies the generation of expressions closely aligned with the borders of the dataset-covered area.

As indicated in Table \ref{tab2}, the majority of the derived expressions demonstrate a satisfactory performance in delineating boundary values, as evidenced by the minimal error committed.  

It is crucial to note that the points were randomly generated over the complete 
surface (volume) of the 2D (3D) figures limits, thus instances laying exactly on the boundaries of the constraints may not be included in the training datasets. Consequently, the resulting mathematical expression can vary significantly based on the specific dataset used and the intricacies of the training process.


\begin{center}
\begin{table}
\caption{Mathematical expression obtained for the 2 features problem  }\label{tab2}
\centering
\begin{tabular}{|c|c|c|}
\hline
Result \# & Expression & Error \\
\hline 
 1 &  $-4.402 \leq X_0$ & 1.45 \% \\ 
 2 &  $18188.7953 \leq - 4002.3703*X_0 + X_1 $  & 1.06 \% \\ 
 3 &  $2.1469 \leq 0.4772*X_0 + X_1  $ & 0.91 \%  \\ 
 4 &  $2.1420 \leq 0.4683*X_0 + X_1 $ & 0.91 \%   \\ 
 5 &  $- 4.6843 \leq - 0.0151*X0 + X_1 $ &  4.68 \%   \\
 6 &  $ 10353.0651 \leq - 2288.4936*X_0 + X_1$ & 1.09 \% \\ 
 7 &   $-4.678 \leq X_1$ & 0.36 \% \\
 8 &  $-4.702 \leq X_1$ & 0.36 \% \\ 
 9 &  $2.0978 \leq 0.4497*X_0 + X_1$ & 0.91 \%  \\ 
 10 &  $ - 0.8634 \leq 1.9262*X_0 + X_1$ & 1.45 \% \\

 \hline 
 11 &  $ 0.5036 \leq 0.2023*X_0 + 0.1505*X_1 - 0.0439*X_2 $   & 1.34 \% \\ 
 12 &  $ - 0.7109 \leq 0.2230*X_0 + 0.0158*X_1 + 2.6140e^{-5}*X_2   $ & 1.65 \% \\ 
 13 &  $ - 0.8440 \leq 0.2247*X_0 + 0.0113*X_1 - 0.0002*X_2  $ & 1.67 \% \\ 
 14 &  $ - 0.8210 \leq - 6.476e^{-5}*X_0 + 0.217*X_1 + 7.475e^{-6}*X_2 $ & 1.47 \% \\ 
 15 &  $ 0.6027 \leq 0.0805*X_0 + 0.2399*X_1 - 4.990e^{-5}*X_2 + $ & 1.37 \% \\

 \hline 
16 &  $ -14.009 \leq -0.4832*X_0 + X_1 $ & 1.36\% \\ 
17 &  $ 14.131 \leq -0.3313*X_0 + X_1 $ & 2.04\% \\ 
18 &  $ - 28.594 \leq -1.9450*X_0 + X_1 $ & 1.36\% \\ 
19 &  $ 20.61634 \leq 1.101*X_0 + X_1 $ & 0.68\% \\ 
20 &  $ -14.975 \leq -1.013*X_0 $ & 1.36\% \\ 
21 &  $ 1.165 \leq 19.252*X_0 + X_1 $ & 0.68\% \\ 
22 &  $ 14.297 \leq 0.28728*X_0 + X_1 $ & 1.36\% \\ 
23 &  $ 2293.265 \leq -174.17*X_0 + X_1 $ & 2.04\% \\ 
24 &  $ -15.398 \leq -0.5665*X_0 + X_1 $ & 0.0\% \\ 
25 &  $ 17.7929 \leq 0.84825*X_0 + X_1 $ & 1.36\% \\ 

\hline 
26 &  $ -5.0622 \leq -0.011*X_0 + X_1 $ & 1.23\% \\ 
27 &  $ -0.139 \leq 1.163*X_0 + X_1 $ & 1.23\% \\ 
28 &  $ 57.826 \leq 11.204*X_0 + X_1 $ & 1.23\% \\ 
29 &  $ 2.353 \leq 0.459*X_0 + X_1 $ & 1.23\% \\ 
30 &  $ 2.330 \leq 0.451*X_0 + X_1 $ & 1.23\% \\ 
31 &  $ 2.348 \leq 0.457*X_0 + X_1 $ & 1.23\% \\ 
32 &  $ 2.364 \leq 0.457*X_0 + X_1 $ & 2.47\% \\ 
33 &  $ 2.3713 \leq 0.459*X_0 + X_1 $ & 2.47\% \\ 
34 &  $ -4.5015 \leq 0.000476*X_0 + X_1 $ & 1.23\% \\ 
35 &  $ 2.351 \leq 0.454*X_0 + X_1 $ & 1.23\% \\ 

\hline
 
\hline
\end{tabular}
\end{table}
\end{center}

The initial ten expressions listed in the table incorporate the masking process, 
resulting in the elimination of relatively small terms from the mathematical 
expressions during the training process. This elimination is enforced after the 
training process. For instance, Result $\# 7$ in the table represents a horizontal 
line, yet the result for the equivalent experiment in Figure \ref{fig1}, produced with no masking, reveals small slopes on the lines (red line), showcasing the impact 
of the masking process on the final expressions.

In Results $\# 11$ to $\# 15$, which correspond to the 3D figure, no masking process was applied (only these results from Table \ref{tab2}). As evident in these expressions, numerous terms could be retained or eliminated based on the relative importance of different features in the final mathematical expressions. The absence of the masking approach allows for a more comprehensive inclusion of terms, potentially enhancing the richness of the expressions. However, it is essential to note that the masking process, as described earlier, contributes to the parsimony of the final mathematical expression, thereby improving readability for users. The choice between using or not using the masking approach depends on the specific goals and interpretability requirements of the application.


Despite the promising results outlined so far, several trends, other than those regularly related to the use of ML (i.e. meta parameters definitions), were observed during the evaluation of the presented approach. These trends were primarily associated with dimensionality and initialization.

In terms of dimensionality, as the number of features increased, unexpected inequalities with an unusual number of features were observed (e.g. See Result \#11 in Table \ref{tab2}). These constraints could arise from the combination of constraints (e.g., combining $-5 \leq X_0$ and $-5 \leq X_1$ as $-10 \leq X_0 + X_1$). A refinement stage, together with an extended number of epochs during training (to improve parameter determination), could be used to secure obtaining more precise inequality expressions. As previously suggested in literature \cite{Beldiceanu_2016}, the incorporation of dominance checks and trivia removal could be considered during this refinement stage. These techniques evaluate constraints between each other or apply specific rules to simplify and eliminate irrelevant constraints, contributing to a more refined and accurate set of inequalities.


The initialization stage involves specifying the weights within the Equation Learner (EQL), and thus initial constraints assumptions to be corrected during training. These initial assumptions could drive gradient values during the training process, and consequently, the direction in which mathematical expressions are explored. 
In the current strategy, weights were randomly initialized for testing the developed approach; however, alternative approaches could be considered to improve searches. For instance, zero initialization for certain values could be enforced, thereby guiding the search towards a reduced number of features involved in the final expression. This thoughtful initialization strategy could potentially contribute to more effective and efficient exploration of the solution space.

\section{Conclusions}\label{conclusion}

In this work, a novel approach for learning constraints from data using DNN based on SR was introduced. The method demonstrated its capability to directly extract linear inequalities from datasets by settling suitable loss functions. The approach was validated on predefined datasets, revealing its effectiveness in approximating the boundaries defined by the constraints. The results, showcase the satisfactory performance of the derived expressions in delineating boundary values, with minimal error committed. The error, quantified as the percentage of points violating the original constraint areas, approached values close to 0, indicating a close alignment with the dataset-covered area borders. The outcomes support the efficacy of the proposed methodology in constraint learning.

The approach has the potential to be extended by incorporating further tailored loss functionalities terms. This extension could facilitate to address diverse challenges and achieve more robust results. Furthermore, extensions such as the use of non-linear primitives in the EQL, together with the exploration of optimal non-linear primitives and constraints, can be easily foreseen to be implemented. As expected, these considerations aim to refine the proposed approach for efficient knowledge extraction from data. 




\begin{credits}
\subsubsection{\ackname}  Science Foundation Ireland under Grant No. 12/RC/2289 for funding the Insight Centre of Data Analytics (which is co-funded under the European Regional Development Fund).

\end{credits}

\end{document}